\documentclass[10pt,twocolumn,letterpaper]{article}

\usepackage[pagenumbers]{cvpr} 

\usepackage[dvipsnames]{xcolor}

\usepackage{url}
\usepackage{amsmath,graphicx}
\usepackage{booktabs}
\usepackage{multirow}
\usepackage{color}
\usepackage{multicol}
\usepackage{algorithm}
\usepackage{algorithmic}
\usepackage{amssymb}

\definecolor{cvprblue}{rgb}{0.21,0.49,0.74}
\usepackage[pagebackref,breaklinks,colorlinks,citecolor=cvprblue]{hyperref}

\def\paperID{1688} 
\def\confName{CVPR}
\def\confYear{2024}

\title{Boosting Audio-visual Zero-shot Learning with Large Language Models}

\author{Haoxing Chen$^{1}$, Yaohui Li$^{2}$, Yan Hong$^{1}$, Zizheng Huang$^{1,2}$, Zhuoer Xu$^{1}$\\  Zhangxuan Gu$^{1}$, Jun Lan$^{1}$, Huijia Zhu$^{1}$, Weiqiang Wang$^{1}$\\
$^{1}$Ant Group, $^{2}$Nanjing University\\
{\tt\small hx.chen@hotmail.com, yaohuili@smail.nju,edu.cn}}

\begin{document}
\maketitle

\begin{abstract}
Audio-visual zero-shot learning aims to recognize unseen classes based on paired audio-visual sequences. Recent methods mainly focus on learning multi-modal features aligned with class names to enhance the generalization ability to unseen categories. However, these approaches ignore the obscure event concepts in class names and may inevitably introduce complex network structures with difficult training objectives.
In this paper, we introduce a straightforward yet efficient framework called KnowleDge-Augmented audio-visual learning (KDA), which aids the model in more effectively learning novel event content by leveraging an external knowledge base.
Specifically, we first propose to utilize the knowledge contained in large language models (LLMs) to generate numerous descriptive sentences that include important distinguishing audio-visual features of event classes, which helps to better understand unseen categories. Furthermore, we propose a knowledge-aware adaptive margin loss to help distinguish similar events, further improving the generalization ability towards unseen classes. Extensive experimental results demonstrate that our proposed KDA can outperform state-of-the-art methods on three popular audio-visual zero-shot learning datasets. Our code will be avaliable at \url{https://github.com/chenhaoxing/KDA}.

\end{abstract}

\section{Introduction}
\begin{figure}[t]
	\centering
	\includegraphics[width=\linewidth]{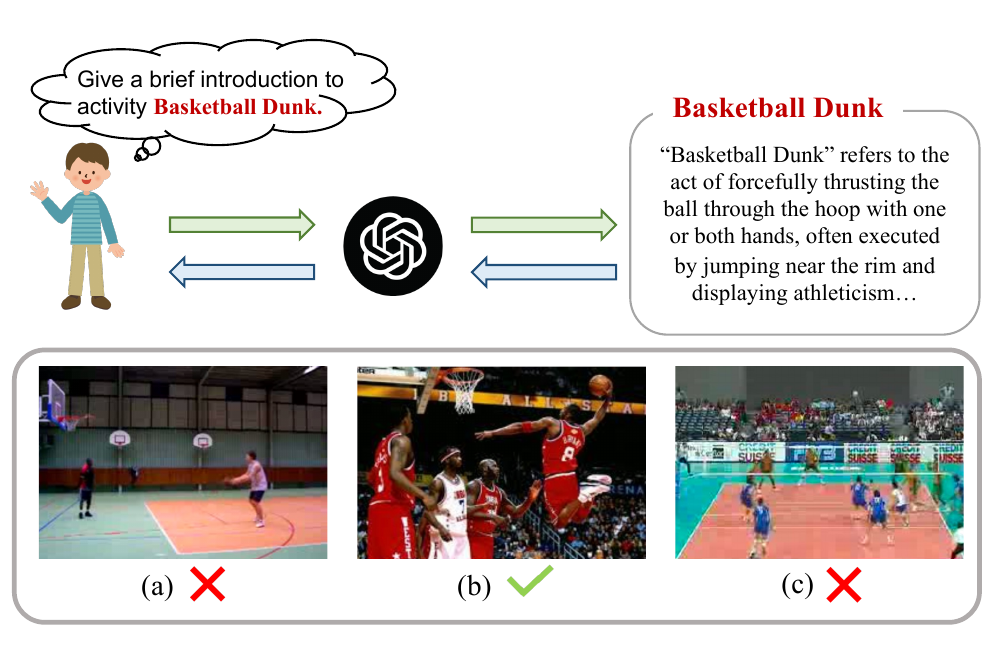}
	\caption{Inspired by the fact that detailed descriptions can help people understand novel concepts and distinguish similar event contents, we propose to improve model generalization ability based on external knowledge.}
	\label{motivation}
\end{figure}

Visual and audio signals frequently occur concurrently in videos. For example, in cinemas, high-quality displays and sound systems are installed to create an immersive experience. 
Humans also make integrated decisions by perceiving through multiple senses. 
In interpersonal communication, individuals often rely on observing facial expressions and body language while listening to others to comprehend their intentions.
Combining audio and visual signals is beneficial for many applications because acoustic features can assist in analyzing emotions, and visual features can help identify speaking objects in a video.
Consequently, there has been growing interest in learning more comprehensive audio-visual representations for multiple tasks~\cite{CASP-Net,egoav,AVFace,MIR-GAN}, such as robotic navigation~\cite{avms,AVLEN}, activity recognition~\cite{CAAV}, highlight detection~\cite{JVAVHD}, and speech recognition~\cite{MIR-GAN}.

Learning audio-visual representations for specific tasks typically requires a large number of annotated samples, which can be time-consuming, expensive, and often impractical in certain applications. Furthermore, the classes present in the existing datasets are limited, making it challenging for models to make accurate predictions when encountering unseen classes. This is due to their lack of necessary knowledge and context to make informed decisions about unfamiliar concepts. Zero-shot recognition~\cite{AVGZSLNet,AVCA,TCaF,HyperbolicAV} task has been proposed to enhance the ability of deep models to recognize unseen classes. In zero-shot recognition tasks, models are required to acquire transferable knowledge to handle data from unfamiliar classes effectively. In this paper, we focus on audio-visual zero-shot learning.

Previous work~\cite{TCaF,HyperbolicAV} has adopted the framework proposed in ACVA~\cite{AVCA} to solve zero-shot video classification tasks using paired audio-visual inputs. Specifically, AVCA learns to map audio-visual features to textual embeddings of class labels to classify samples from unseen classes. However, due to the limitations of text label representations with coarse granularity and weak representation ability, directly mapping audio-visual features to label feature space is not efficient and may introduce understanding bias.

\begin{figure}[t]
\centering
\includegraphics[width=0.99\linewidth]{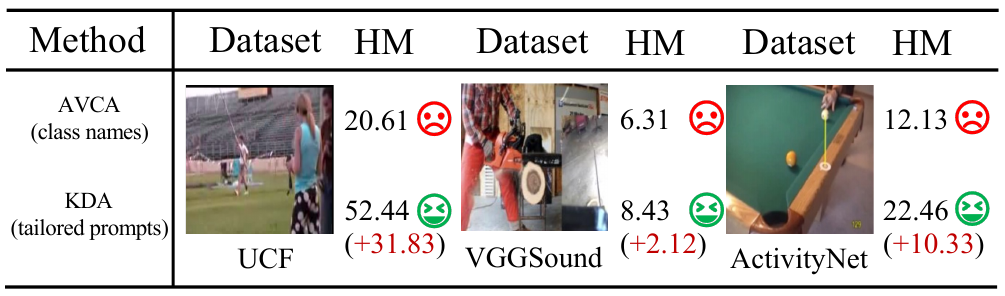}
	\caption{We perform audio-visual zero-shot classification experiments on three benchmark datasets. We can find that textual descriptions with richer knowledge improve the generalization ability of models.}
	\label{fig2}
\end{figure}

Inspired by how humans utilize prior knowledge to learn novel visual concepts, in this paper, we propose a novel KnowleDge-Augmented  audio-visual learning (KDA) framework, which aids the model in more effectively learning novel event content by leveraging an external knowldge base. Specifically, we utilize large language models~\cite{brown2020language,touvron2023llama} (LLMs) as external knowledge bases to generate detailed descriptions of event concepts.
As a motivating example, we can see in Figure~\ref{motivation} that it can be difficult to distinguish between several similar events that have not been encountered before (e.g., Basketball Dunk V.S. Basketball Dunk).
However, once the model acquires detailed descriptions of events from the external knowledge base, it becomes easier to recognize the events and establish the corresponding relationships between different categories.
To better utilize the description of event concepts, within the KDA framework, we map audio-visual features and textual knowledge features to a common space and use alignment loss to ensure intra-class compactness, meaning that samples of the same category are as close as possible to their corresponding event descriptions. Additionally, to enhance inter-class separability of features, we propose a knowledge-aware adaptive margin loss. By adding adaptive margin to the classification loss, the knowledge-aware adaptive margin loss effectively separates different categories. This adaptive margin is generated based on the knowledge similarity between each pair of categories. As is shown in Figure~\ref{fig2}, experiment results verify that more detailed text improves performance of model on several event datasets. 

Our main contributions are summarized as follows:
\begin{itemize}
    \item We propose a novel audio-visual zero-shot learning framework by leveraging knowledge from large language models, which greatly improves the generalization ability on unseen event classes.
    \item We propose a knowledge-aware adaptive margin loss to further separate different classes in the common embedding space according to their description similarities.
    \item Extensive experiments demonstrate that, the proposed KDA outperform existing models. And we perform a detailed analysis of the different proposed distribution alignment methods, demonstrating the benefits of our proposed model architecture.
\end{itemize}

\section{Related Works}
\subsection{Audio-visual Zero-shot Learning}
Parida et al.~\cite{CJME} introduced the Audio-Visual Zero-Shot Learning (AVZSL) task and proposed the Coordinated Joint Multimodal Embedding (CJME) model, which aims to map video, audio, and text into a shared feature space for comparison. To ensure that video or audio features are closely aligned with their corresponding class features, the triplet loss was employed. This loss function encourages the model to learn a feature representation where instances of the same class are closer to each other than instances of different classes, thereby facilitating the zero-shot learning process.
Mazumder et al.~\cite{AVGZSLNet} proposed the Audio-Visual Generalized Zero-shot Learning Network (AVGZSLNet) to address the challenges of audio-visual zero-shot learning. The AVGZSLNet incorporates a module that reconstructs text features using visual and audio features, which helps bridge the gap between different modalities and improves the model ability to generalize to unseen classes.
The Audio-Visual Cross-Attention (AVCA) framework~\cite{AVCA} was specifically designed to facilitate the exchange of information between video and audio representations. This framework enables the generation of informative representations that contribute to achieving state-of-the-art performance in audio-visual zero-shot classification tasks.
Mercea et al.~\cite{TCaF} proposed a multi-modal and Temporal Cross-attention Framework (TCaF), which is designed to leverage the temporal dynamics of audio-visual data and the cross-modal relationships between different sensory inputs. This framework aims to improve the model ability to understand and interpret complex audio-visual scenes.
Hong et al.~\cite{HyperbolicAV} introduced a novel approach that incorporates hyperbolic learning with a hierarchical structure into the AVZSL task. This approach achieved promising results by leveraging the geometric properties of hyperbolic spaces to model the relationships between different classes and modalities.
Unlike previous methods, we utilize large language models to generate more detailed event descriptions, which promotes the model understanding of events. Specifically, we map audio-visual features and event descriptions to a common space and ensure intra-class compactness of features through alignment loss, and inter-class separability through adaptive margin loss.

\subsection{Margin Loss in Visual Recognition}
Softmax loss, a widely adopted technique in the realm of machine learning, plays a pivotal role in the training of Convolutional Neural Networks (CNNs) for the purpose of extracting features that are crucial for object recognition tasks. The weights of the final fully connected layer in a classification CNN, when trained using Softmax loss, have been observed to bear a conceptual resemblance to the centroid of each class. This observation has led to the development of various margin loss approaches, such as Arcface, SphereFace, and CosFace, which are designed to enhance the discriminative power of the extracted features.
Liu et al.\cite{SphereFace} introduced the concept of angular margin, which, while innovative, necessitates approximate computation, a factor that can lead to instability in the training process of the network. In contrast, Wang et al.\cite{CosFace} proposed a method that directly incorporates the cosine margin into the objective function, achieving superior results compared to the approach taken by Liu et al.\cite{SphereFace}. Furthermore, Deng et al.\cite{Arcface} introduced the corner margin loss, which serves to further augment the discriminative capabilities of the feature space.
Despite the impressive results achieved by these margin losses in the field of visual understanding tasks, they are not ideally suited for AVZSL, a multi-modal task that presents a unique challenge due to the absence of samples for novel classes. To address this issue, we propose a novel principle for generating a knowledge-guided adaptive margin. Our proposed knowledge-aware adaptive margin loss is designed to be trained alongside AVZSL methods, enabling the model to effectively align audio-visual representations with knowledge representations while maintaining separability between different classes. This alignment results in improved generalization performance on unseen categories, thereby enhancing the overall effectiveness of the model in the context of AVZSL.

\begin{figure*}[t]
	\centering	\includegraphics[width=0.99\linewidth]{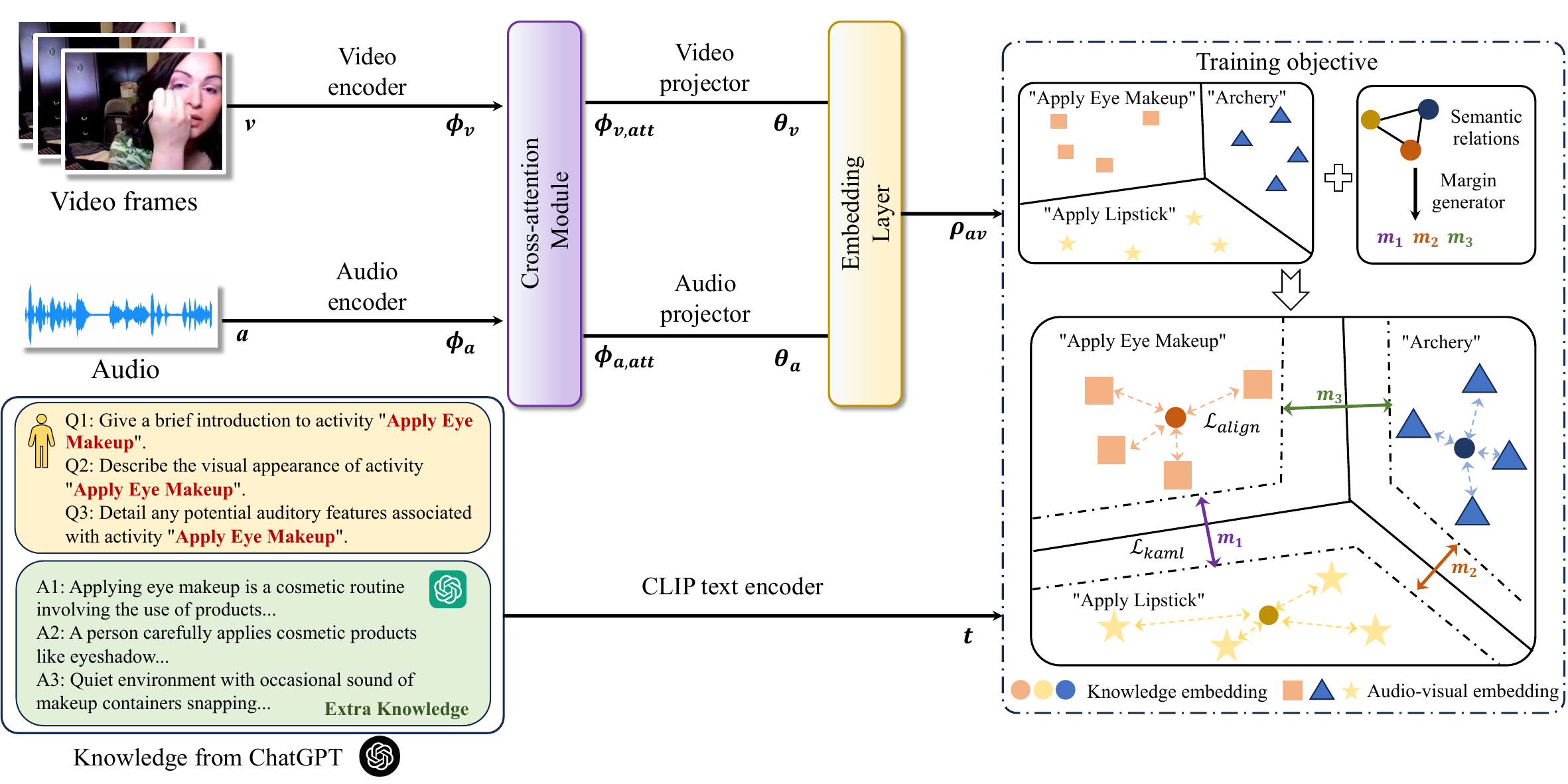}
	\caption{Overview of our proposed KnowleDge-Augmented  audio-visual learning (KDA). KDA takes the audio and visual features extracted from the video data as input, and obtains multi-modal audio-visual features $\rho_{av}$ for classification through the cross-attention module and embedding layer. The knowledge description is obtained through the interpretation of the event name by LLMs, and then the knowledge representation $t$ is obtained by using the CLIP text encoder. We use alignment loss $\mathcal{L}_{align}$ to promote intra-class compactness learning, utilizing knowledge-aware adaptive loss $\mathcal{L}_{kaml}$ to enhance inter-class separability learning.
}
	\label{KAML}
\end{figure*}

\subsection{Large Language Models for Downstream Tasks}
In recent literature, there has been a surge of interest in the application of LLMs across various domains, with researchers exploring different ways to harness the power of these models for tasks such as image captioning, commonsense reasoning, and multimodal understanding. Santurkar et al.~\cite{santurkar2022caption} introduced CLIP$_{\rm s}$, a method that leverages LLMs to interpret and augment existing image captions, which are then used to enhance the performance of the CLIP framework. Liu et al.~\cite{KP4cr} proposed a novel approach called generated knowledge prompting, which involves eliciting and integrating knowledge from GPT-3 to improve performance on commonsense reasoning tasks.
Su et al.~\cite{su2022language} presented MAGIC, a groundbreaking decoding scheme that integrates visual controls into the generation process of a LLM. MAGIC is a training-free framework that allows LLMs to tackle complex multimodal tasks in a zero-shot manner without compromising decoding speed. Similarly, Yang et al.~\cite{yang2022empirical} utilized GPT-3 in conjunction with text descriptions of images for the Visual Question Answering (VQA) task, demonstrating the versatility of LLMs in multimodal applications.
Building upon these advancements, our work specifically focuses on leveraging LLMs to enhance the zero-shot generalization capabilities of models. We aim to tap into the vast knowledge stored within LLMs to improve the ability of models to recognize and understand unseen event classes. By doing so, we aim to enhance the generalization performance of models in the context of audio-visual zero-shot learning, where the models are required to recognize and classify objects or events based on both audio and visual cues, even when no direct examples of the target classes are provided during training. Our approach is designed to align with the principles of knowledge-aware adaptive margin loss, ensuring that the model can effectively align audio-visual representations with knowledge representations, leading to improved performance on unseen classes.

\section{Knowledge-augmented audio-visual learning}
\subsection{Preliminary}
We first introduce the concept of audio-visual zero-shot learning (AVZSL). The objective of AVZSL is to recognize and classify classes that were not present during the training phase, specifically, identifying videos from unseen classes. This task poses substantial challenges, as it necessitates the model to generalize from previously encountered classes to categorize samples from novel, unseen classes effectively. In the generalized zero-shot learning (GZSL), the complexity is further heightened as the test set comprises samples from both unseen and previously encountered classes, rendering the task more representative of real-world scenarios and thereby more challenging.

Formally, let the training set be defined as $S=\{(a^s_i, v^s_i, w^s_i, y^s_i)\}_{i=1}^{N}$, which exclusively consists of samples from known classes.  Here, $a^s_i$, $v^s_i$, and $w^s_i$ denote the audio, visual, and class-level text embeddings, respectively, and $y^s_i$ corresponds to the ground-truth label.
The primary objective is to construct a predictive model $f$: $f(v^s_i, a_s^i)\longmapsto w_s^i$, which is expected to generalize its application to samples from unseen classes, thereby enabling it to operate such that $f(v^u_i, a^u_i)\longmapsto w^u_i$ post-training. Here, $U=\{(a^u_i, v^u_i, w^u_i, y^u_i)\}_{i=1}^{M}$ symbolizes the test sample set derived from these unseen classes.
The main objective is to develop an effective model that can successfully transfer knowledge learned from seen classes to recognize and classify samples from unseen classes.

\subsection{Tailored Prompts through Language Models}
In the following, we describe the details of our proposed TaPL (see Figure \ref{KAML}). 
The TaPL involves a two-phase process: firstly, it crafts tailored prompts for every class within a specified dataset; secondly, it applies these prompts to execute audio-visual zero-shot classification.

\noindent
\textbf{Generating tailored prompts.} 
Previous methods~\cite{AVCA,AVGZSLNet} employed class names for textual representation and harnessed audio-visual attributes for alignment (e.g., "Skateboarding", "Skiing"). Nonetheless, the informational content inherent in these class names is insufficient, impeding the effective correspondence between audio-visual attributes and textual features, particularly when differentiating events that exhibit similarity in their categorical designations.
Inspired by how humans utilize knowledge bases to gain detailed understanding of new visual concepts, we propose using LLMs to generate more detailed descriptions for event concepts. As shown in Figure \ref{motivation}, humans often struggle to imagine the performance of uncommon event classes, and it becomes even more challenging to recognize them when encountering event nouns for the first time. Additionally, it is also easy to confuse visually similar event classes, such as "playing basketball" and "slam dunk". However, once we provide more detailed explanations of event nouns through a knowledge base, it becomes much easier to identify the correspondences between different videos and different event classes.

Specifically, we employ two distinct prompt methods to obtain feature representations for each event category: one is based on handcrafted templates, and the other is generated by LLMs. For datasets dominated by natural sounds, such as VGGSound~\cite{Vggsound}, and those dominated by human actions, such as UCF~\cite{UCF101} and ActivityNet~\cite{Activitynet}, we have designed different prompt templates.

(1) Handcrafted templates: We construct a basic description for UCF and ActivityNet using the format "A video of a person doing \{event name\}." For VGGSound, we build a basic description using the format "A video of event \{event name\}." This template is designed to provide a simple and straightforward description of the event, which is easily understandable by both humans and machines.

(2) LLM-generated: We apply multiple prompts to LLMs across each dataset to procure representations of knowledge. For UCF and ActivityNet, we utilize the following three LLM prompts:
\begin{itemize}
    \item Give a brief introduction to activity \{event name\}.
    \item Describe the visual appearance of activity \{event name\}.
    \item Detail any potential auditory features associated with activity \{event name\}.
\end{itemize}     
And for VGGSound, we utilize the following three LLM prompts:
\begin{itemize}
    \item Give a brief introduction to sound event \{event name\}.
    \item Describe the visual representation of a scenario characterized by \{event name\} event sound.
    \item Detail auditory features associated with sound event \{event name\}.
\end{itemize} 
These prompts are designed to elicit more detailed and nuanced descriptions of the event, which can provide additional context and information that can be useful for understanding and recognizing the event.

Following this, we encode both the base and LLM generated descriptions utilizing the CLIP text encoder, deriving the knowledge representation $t$. Note that the basic description and three types of generated descriptions are separately fed into CLIP, after which we average the obtained features to get $t$. Thus, the composition of the data set $S$ and $U$ becomes $S=\{(a^s_i, v^s_i, t^s_i, y^s_i)\}_{i=1}^{N}$ and $U={(a^u_i, v^u_i, t^u_i, y^u_i)}_{i=1}^{M}$ respectively. This enhancement allows for more informative and detailed knowledge representation, enabling the model to better understand and discriminate between different event classes.

\noindent
\textbf{Utilizing tailored prompts.}
Instead of using separate visual (audio) and text feature spaces for classification~\cite{AVCA,HyperbolicAV}, we map audio-visual features and knowledge representations into a common space and then classify them. 
To achieve this, we first extract uni-modal features, denoted as $\theta_a$ and $\theta_v$, using encoder blocks $\mathbf{A}_{enc}$ and $\mathbf{V}_{enc}$, a cross-attention module $\mathbf{F}_c$, and projectors $\mathbf{A}_{proj}$ and $\mathbf{V}_{proj}$, respectively.
Subsequently, we employ the embedding layer, denoted as $\mathbf{E}_{av}$, to map the concatenated multi-modal feature into the textual space, thereby acquiring the multi-modal feature $\rho_{av}$.

To ensure that the audio-visual features of samples from the same class are aligned as closely as possible with their corresponding knowledge features, we minimize the Euclidean distance between these feature sets. This objective is defined by the following formula:

\begin{equation}
     \mathcal{L}_{align} = ||\rho_{av}-t||^2_2.
\end{equation}

The alignment loss $\mathcal{L}_{align}$ ensures that the feature representations $\rho_{av}$ and $t$ belonging to the same class are as close as possible in the common feature space. However, it does not guarantee that $\rho_{av}$ and $t$ from different classes are sufficiently separated. For the classification of audio-visual samples, we compute the dot product of $\rho_{av}$ with the knowledge representation of each class $k$, thereby deriving the class logits $s^k = (t^k)^\top \rho_{av}$.
These scores serve as a measure of similarity, indicating the degree of alignment between the audio-visual features and the corresponding knowledge features of each class. To encourage the audio-visual representation to achieve higher scores in alignment with the corresponding knowledge representation and lower scores with other knowledge representations, we apply cross-entropy loss, which penalizes the discrepancies between the predicted class probabilities and the actual labels:
\begin{equation}
\mathcal{L}_{cls} = -\frac{1}{N} \sum_{{a,v,y}\in S} {\rm log}\,\frac{e^{s^y}}{\sum\limits_{k\in C_t} e^{s^k}},
\end{equation}
where $C_t$ is the seen classes in $S$ and $N$ denotes the sample number of training set. 

However, the original Cross entropy loss is not very effective at separating similar classes. To enhance the separation of similar classes within the feature embedding space, the margin between two classes should be adaptively determined. Specifically, for classes that are more similar, the margin should be greater than for those that are less similar. To address this, we propose a novel knowledge-aware adaptive margin loss $\mathcal{L}_{kaml}$, which leverages the semantic similarities between class knowledge representations to adjust the margin. Specifically, for each pair of classes $i$ and $j$, we utilize their corresponding knowledge representations $t^i$ and $t^j$ to compute the margin $m_{i,j}$ using the following formulation::
\begin{equation}
    m_{i,j} = \alpha \cdot {\rm cos}(t^i, t^j) + \beta,
\end{equation}
where ${\rm cos(.,.)}$ denotes the cosine similarity, $\alpha$ and $\beta$ denotes the scale and bias parameters respectively. By incorporating the knowledge-aware margin into the classification loss, we derive the $\mathcal{L}_{kaml}$:
\begin{equation}
    \mathcal{L}_{kaml} = -\frac{1}{N} \sum_{{a,v,y}\in S} {\rm log}\,\frac{e^{s^y}}{e^{s^y}+\sum\limits_{k\in C_t \backslash \{y\} } e^{s^k+m}}.
\end{equation}
Our proposed knowledge-aware adaptive margin loss leverages the knowledge similarity between classes to augment the separability of samples from similar classes within the shared embedding space. By modulating the classification margin according to inter-class knowledge proximities, our approach fosters a more distinct embedding landscape. This improved separability and discriminability of features contribute to enhanced recognition of previously unseen test classes, thereby enabling the model to effectively generalize to novel and unseen classes.

\begin{table*}[t]
    \centering
    \caption{Experimental results of audio-visual zero-shot learning on three datasets (main feature). The mean class accuracy for GZSL is reported on the seen (S) and unseen (U) test classes, and their harmonic mean (HM). For the ZSL performance, only the test subset of unseen classes is considered.}
    \resizebox{\textwidth}{!}{
    \begin{tabular}{cccccccccccccc}
    \toprule
       \multirow{2}{*}{Model} & \multirow{2}{*}{Venue} & \multicolumn{4}{c}{UCF-GZSL} & \multicolumn{4}{c}{VGGSound-GZSL} & \multicolumn{4}{c}{ActivityNet-GZSL}\\ \cline{3-14}&&S&U&HM&ZSL&S&U&HM&ZSL&S&U&HM&ZSL\\
       \midrule
       DEVISE~\cite{DEVISE}&NeurIPS'13&55.59&14.94&23.56&16.09&36.22 &1.07&2.08& 5.59&3.45&8.53&4.91&8.53\\
        ALE~\cite{ALE}&T-PAMI'15&57.59&14.89&23.66&16.32&0.28&5.48& 0.53& 5.48&2.63&7.87&3.94& 7.90\\
        SJE~\cite{SJE}&CVPR'20&63.10&16.77&26.50&18.93&48.33& 1.10 &2.15 &4.06&4.61& 7.04& 5.57& 7.08\\
         f-VAEGAN-D2~\cite{fVAEGAND2}&CVPR'19&17.29&8.47 & 11.37 &11.11&12.77 &0.95& 1.77 &1.91&4.36 &2.14& 2.87 &2.40\\
         CJME~\cite{CJME}&WACV'20&26.04& 8.21&12.48 & 8.29&8.69& 4.78& 6.17& 5.16& 5.55& 4.75& 5.12& 5.84\\
         AVGZSLNet~\cite{AVGZSLNet}&WACV'21&52.52& 10.90 &18.05& 13.65&18.05 &3.48& 5.83& 5.28&8.93& 5.04& 6.44& 5.40\\
         APN~\cite{APN}&IJCV'22&28.46 &16.16&20.61 &16.44&7.48& 3.88 &5.11 &4.49 &9.84 &5.76& 7.27& 6.34\\
         AVCA~\cite{AVCA}&CVPR'22&51.53&18.43& 27.15 & 20.01&14.90& 4.00& 6.31& 6.00 &24.86& 8.02& 12.13 &9.13\\
         TCaF~\cite{TCaF}&ECCV'22&58.60&21.74& 31.72& 24.81 &9.64& 5.91& 7.33& 6.06& 18.70& 7.50 &10.71& 7.91\\
         VIB-GZSL~\cite{VIB}&ICME'23&90.35&21.41&34.62& 22.49&18.42& 6.00& 9.05& 6.41& 22.12& 8.94& 12.73 &9.29\\
         ACFS~\cite{acfs}&IJCNN'23&54.87& 16.49& 25.36& 22.37&15.20& 5.13& 7.67& 6.20& 29.00& 9.13& 13.89& 11.18\\
         Hyper-multiple~\cite{HyperbolicAV}&ICCV'23&63.08& 19.10& 29.32& 22.24&15.02& 6.75& 9.32& 7.97& 23.38& 8.67 &12.65 &9.50\\
         MDFT~\cite{MDFT}&ACM MM'23&48.79&23.11&31.36&31.53&16.14&5.97&8.72&7.13&18.32&10.55&13.39&12.55\\
        \midrule
        KDA&Ours&91.99&36.67&\textbf{52.44}&\textbf{39.06}&24.10& 7.22&\textbf{11.11}&\textbf{8.43}&37.95&15.95&\textbf{22.46}&\textbf{17.23}\\
        \bottomrule
    \end{tabular}}
    \label{table1}
\end{table*}

\noindent
\textbf{Inference.} 
During test time, we determine the class prediction $c$ by identifying the knowledge representation that is closest to the multi-modal representation $\rho_{av}$. This is achieved by measuring the distance or similarity between $\rho_{av}$ and each knowledge representation, and selecting the class whose knowledge representation has the smallest distance or highest similarity to $\rho_{av}$. The predicted class $c$ corresponds to the class with the most similar knowledge representation to $\rho_{av}$:
\begin{equation}
    c = \mathop{\arg\min}\limits_{i}(||t^i-\rho_{av}||_2).
\end{equation}

\subsection{Optimization}
Our full model optimizes the cross-attention module, encoders, projectors, embedding layer, and decoders simultaneously. This is achieved by minimizing the following objective function:
\begin{equation}
    \mathcal{L}_{\rm KDA} = \mathcal{L}_{kaml} + \lambda \cdot \mathcal{L}_{align},
\end{equation}
where $\lambda$ is the weight that control the importance of the alignment loss. It is worth noting that our model consistently achieves significant results across all datasets, demonstrating its robustness and ease of training. These consistent and impressive results validate the effectiveness and reliability of our proposed method.

\begin{table*}[t]
    \centering
    \caption{Experimental results of audio-visual zero-shot learning on three datasets (classification feature). The mean class accuracy for GZSL is reported on the seen (S) and unseen (U) test classes, and their harmonic mean (HM). For the ZSL performance, only the test subset of unseen classes is considered.}
    \resizebox{\textwidth}{!}{
    \begin{tabular}{cccccccccccccc}
    \toprule
       \multirow{2}{*}{Model} & \multirow{2}{*}{Venue} & \multicolumn{4}{c}{UCF-GZSL$^{cls}$} & \multicolumn{4}{c}{VGGSound-GZSL$^{cls}$} & \multicolumn{4}{c}{ActivityNet-GZSL$^{cls}$}\\ \cline{3-14}&&S&U&HM&ZSL&S&U&HM&ZSL&S&U&HM&ZSL\\
       \midrule
       DEVISE~\cite{DEVISE}&NeurIPS'13&29.58& 34.80& 31.98& 35.48 &29.96& 1.94& 3.64& 4.72& 0.17& 5.84& 0.33& 5.84\\
        SJE~\cite{SJE}&CVPR'20&19.39& 32.47& 24.28& 32.47& 16.94& 2.72& 4.69& 3.22 &37.92 &1.22& 2.35& 4.35\\
         CJME~\cite{CJME}&WACV'20&33.89 &24.82& 28.65& 29.01& 10.86 &2.22 &3.68& 3.72& 10.75& 5.55 &7.32&6.29\\
         AVGZSLNet~\cite{AVGZSLNet}&WACV'21&74.79 &24.15& 36.51& 31.51& 15.02 &3.19 &5.26& 4.81 &13.70& 5.96& 8.30& 6.39\\
         APN~\cite{APN}&IJCV'22&13.54& 28.44& 18.35& 29.69& 6.46& 6.13& 6.29& 6.50& 3.79& 3.39 &3.58 &3.97\\
         AVCA~\cite{AVCA}&CVPR'22&63.15& 30.72& 41.34& 37.72& 12.63& 6.19& 8.31& 6.91& 16.77& 7.04& 9.92& 7.58\\
         TCaF~\cite{TCaF}&ECCV'22&67.14& 40.83& 50.78& 44.64& 12.63 &6.72 &8.77 &7.41& 30.12& 7.65& 12.20 &7.96\\
         ACFS~\cite{acfs}&IJCNN'23&54.57&36.94&44.06&41.55& 12.87&5.22& 7.43& 6.03& 14.41& 8.91&11.01&9.15\\
         Hyper-multiple~\cite{HyperbolicAV}&ICCV'23&74.26& 35.79& 48.30&\textbf{52.11}& 15.62& 6.00& 8.67& 7.31& 36.98& 9.60& 15.25& 10.39\\
        \midrule
        KDA&Ours&80.03&38.55&\textbf{52.04}&41.48&14.64&6.63&\textbf{9.12}&\textbf{8.00}&43.98&14.98&\textbf{22.35}&\textbf{15.98}
        \\
        \bottomrule
    \end{tabular}}
    \label{table2}
\end{table*}

\section{Experiments}
In this section, we validate the effectiveness of KDA and analyze its components empirically. We first detail of our experimental settings, then present our experimental results and compare KDA with previous state-of-the-art models. Finally, we present an ablation study which shows the benefits of using our proposed methods.

\subsection{Experimental Setup}

\textbf{Datasets}. We perform extensive experimentsto verify the effectiveness of our method on three audio-visual zero-shot learning datasets: VGGSound-GZSL~\cite{AVCA}, UCF-GZSL~\cite{AVCA}, and ActivityNet-GZSL~\cite{AVCA}. 
VGGSound-GZSL, UCF-GZSL, and ActivityNet-GZSL are modified versions of existing audio-visual and event recognition datasets~\cite{Vggsound,UCF101,Activitynet}. VGGSound-GZSL consists of 42 seen and 234 unseen classes. UCF-GZSL includes 21 seen and 30 unseen classes. ActivityNet-GZSL is based on the event recognition dataset ActivityNet and includes 200 classes, with 99 seen classes and 101 unseen classes.
We perform two types of experiments depending on the network used to extract the features, i.e., main features and classification features.
Specifically, we use the self-supervised SeLaVi~\cite{Asano_Patrick_Rupprecht_Vedaldi_2020} to obtain main features, and use C3D~\cite{C3D} and VGGish~\cite{Vggish} to obtain classification features.
To distinguish between the two settings, we add superscript ${cls}$ to the data set names, e.g., UCF-GZSL$^{cls}$.

\noindent
\textbf{Training Details}.
We use the Adam optimizer~\cite{adam} to train all our models. 
Regarding hyperparameters, we set the $\lambda$ to 10.0 for UCF-GZSL/UCF-GZSL$^{cls}$/Activity-GZSL/Activity-GZSL$^{cls}$ and 5.0 for VGGSound-GZSL/VGGSound-GZSL$^{cls}$.
The optimizer has running average coefficients $\beta_1=0.5$, $\beta_1=0.999$, and an initial learning rate of 0.001. We reduce the learning rate by a factor of 0.1 when the GZSL performance plateaus with a patience of 4 epochs. For all datasets, we set the batch size to 2048.
To prevent overfitting, we used dropout ratios of 0.2/0.3 for the encoder and projector on UCF-GZSL/UCF-GZSL$^{cls}$; 0.2/0.2 for Activity-GZSL/Activity-GZSL$^{cls}$; and 0/0 for VGGSound-GZSL/VGGSound-GZSL$^{cls}$.
All experiments are conducted on a single Tesla A100 GPU.

\noindent
\textbf{Evaluation}.
Following~\cite{AVCA}, we use the metrics S, U, ZSL and ${\rm HM = \frac{2US}{U+S}}$ to evaluate the performance of the model in the seen and unseen classes. Specifically, ZSL is obtained by considering only a subset of test samples obtained from the unseen test classes, and HM is the harmonic mean of averaged performance on the unseen and seen classes. All these metrics are evaluated on "main feature" and "cls feature".

\noindent
\textbf{Compared Methods.}
We compare our KDA to five ZSL and eight current state-of-the-art audio-visual ZSL frameworks. The ZSL approaches
include ALE~\cite{ALE}, SJE~\cite{SJE}, DEVISE~\cite{DEVISE}, APN\cite{APN}, and
f-VAEGAN-D2~\cite{fVAEGAND2}. The first four approaches are image-based ones, and f-VAEGAN-D2 is a generative method for
ZSL. For these approaches, we concatenate image and audio features as input instead of using only image features.
The compared audio-visual GZSL approaches are CJME~\cite{CJME},
AVGZSLNet~\cite{AVGZSLNet}, AVCA~\cite{AVCA} TCaF~\cite{TCaF}, VIB-GZSL~\cite{VIB}, ACFS~\cite{acfs}, Hyper-multiple~\cite{HyperbolicAV} and MDFT~\cite{MDFT}.

\begin{figure*}[t]
	\centering\includegraphics[width=0.999\linewidth]{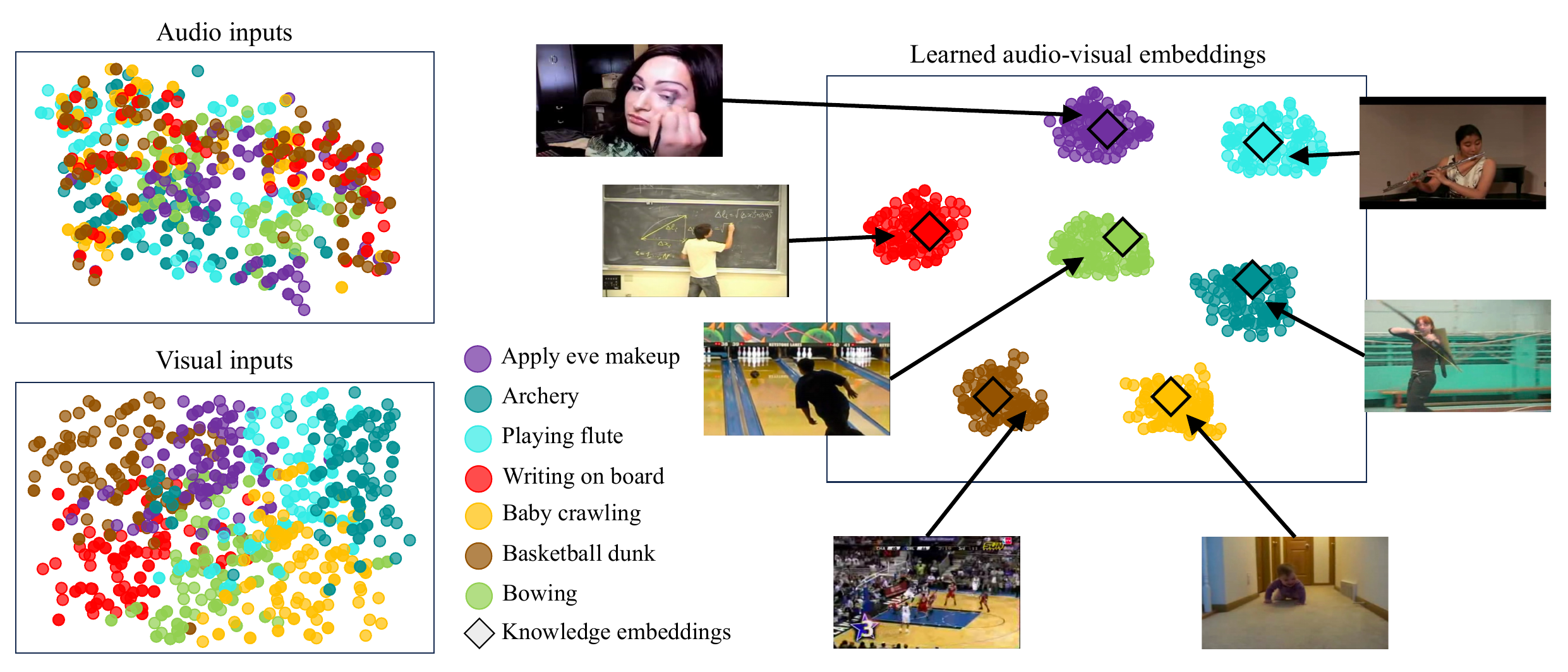}
	\caption{t-SNE visualisation for five seen and two unseen test classes from the UCF-GZSL dataset, showing audio and visual input embeddings extracted with SeLaVi~\cite{Asano_Patrick_Rupprecht_Vedaldi_2020}, and learned audio-visual embeddings in the common space. Knowledge embeddings are visualised with a square. KDA facilitates pulling together features from the same parent class while pushing away features belonging to different parent classes.}
	\label{tsne}
\end{figure*}

\subsection{Experimental Results}

\textbf{Comparison with state-of-the-art.}
To validate the effectiveness of our model, we have compared it with the current state-of-the-art audio-visual ZSL methods on
three benchmark datasets. The main results are presented in Table~\ref{table1} and Table~\ref{table2}. 

For the main feature setting, KDA achieves state-of-the-art performance in all cases.
For example, on UCF-GZSL, KDA achieves a harmonic mean (HM) of 52.44\% and a zero-shot learning (ZSL) performance of 39.06\%, compared to a HM of 29.32\% and a ZSL performance of 22.24\% for the most recent ICCV'23 method Hyper-multiple~\cite{HyperbolicAV}.
On VGGSound-GZSL, KDA obtains a HM of 11.11\% for GZSL and a ZSL performance of 8.43\% compared to 7.33\% HM and 6.06\% ZSL for TCaF. On ActivityNet-GZSL, KDA outperforms AVCA, with an HM/ZSL performance of 19.67\%/14.00\% compared to 12.13\%/9.13\%.

And for the classification feature setting, KDA also achieves state-of-the-art performance in all datasets.
For ActivityNet-GZSL$^{cls}$, our proposed KDA is significantly better than its strongest competitor TCaF, with a HM of 22.35\% compared to 12.20\% Similar patterns are exhibited for the UCF-GZSL$^{cls}$ and ActivityNet-GZSL$^{cls}$ datasets.

In our study, the significant improvement observed provides solid evidence for the effectiveness of detailed knowledge descriptions in enhancing the generalization capability of audio-visual zero-shot learning. By generating meticulous descriptions of event categories and integrating them with the proposed KDA framework for in-depth learning, our model can more accurately understand events, leading to superior generalization performance.

\noindent
\textbf{Qualitative results.}
We present a qualitative analysis of the learnt audio-visual embeddings in Figure~\ref{tsne}. For this, we conduct t-SNE visualization~\cite{hinton_tsne} of audio-visual/knowledge embeddings mapped by our KDA on 7 classes from UCF-GZSL test set.
As shown in Figure~\ref{tsne}, audio and visual input features are poorly clustered, while audio-visual features are well clustered for seen and unseen classes with clear boundaries. This observation shows that the audio-visual features learned by our KDA improve over the clustering of input features for both, seen and unseen classes. 
In addition, knowledge representations lie inside the corresponding audio-visual clusters. This confirms that the learned audio-visual embeddings are mapped close to the corresponding knowledge representation, indicating that our audio-visual embeddings and knowledge embeddings have an excellent distribution alignment.

\begin{table*}[t]
    \centering
    \caption{Ablation study on KDA. The mean class accuracy for GZSL is reported on the seen (S) and unseen (U) test classes, and their harmonic mean (HM). For the ZSL performance, only the test subset of unseen classes is considered.}
    \resizebox{\textwidth}{!}{
    \begin{tabular}{ccccccccccccc}
    \toprule
       \multirow{2}{*}{Model} &  \multicolumn{4}{c}{UCF-GZSL} & \multicolumn{4}{c}{VGGSound-GZSL} & \multicolumn{4}{c}{ActivityNet-GZSL}\\ \cline{2-13}&S&U&HM&ZSL&S&U&HM&ZSL&S&U&HM&ZSL\\
       \midrule
       baseline&96.35&23.18&37.37&28.47&23.85&5.71&9.21&6.17&26.40&13.98&18.28&15.59\\
       w/o handcraft descriptions&91.85&35.40&51.11&37.43&26.22&5.97&9.73&6.41&42.53&  14.22&21.32&15.56\\
       w/o LLM-generated descriptions& 92.66 &29.94&45.26&32.33&24.00&5.97&9.56&6.70&32.15&11.22&16.63&12.04 \\
       w/o $\mathcal{L}_{align}$&45.19&17.17& 24.88&20.10&8.58&4.29& 5.72&5.10&2.75&3.65&3.13&4.80\\
       w/o  $\mathcal{L}_{kaml}$&91.32&34.75&50.34&38.92&25.62& 6.32&10.14&6.70&29.62& 11.48&16.55&12.00\\
        \midrule
        KDA&91.99&36.67&\textbf{52.44}&\textbf{39.06}&24.10&7.22&\textbf{11.11}&\textbf{8.43}&37.95&15.95&\textbf{22.46}&\textbf{17.23}\\
        \bottomrule
    \end{tabular}}
    \label{ablation}
\end{table*}

\subsection{Ablation Study}
Here, we analyze the different components of our proposed KDA. First, we compare the performance of the model when trained using different loss functions and different text representations. Then, we study the impact of $\lambda$ used in KDA on (G)ZSL performance. Finally, we investigate the effect of different text encoders and various modal features on performance.

\noindent
\textbf{Ablation study on key components.}
In our framework, descriptions generated by LLMs, alignment loss, and knowledge-aware adaptive margin loss are the key to improving performance.  
To verify the impact of these components, we designed a series of ablation experiments to analyze their contributions to the overall performance of the model, particularly focusing on the recognition of unseen categories. 
Specifically, we conducted experiments by removing each part from KDA and observing the changes in overall performance. The experimental results are summarized in Table~\ref{ablation}.
Note that the baseline utilizes class names extracted by CLIP as text representations and is trained solely using the align loss and the cross-entropy loss without margin.
We found that compared to the baseline, our KDA achieved a 40.3\%/20.6\%/22.9\% improvement in HM on the UCF-GZSL/VGGSound-GZSL/ActivityNet-GZSL datasets.
When LLM-generated descriptions were not used, the HM of KDA on UCF-GZSL, VGGSound-GZSL, and ActivityNet-GZSL dropped by 13.7\%, 14.0\%, and 26.0\% respectively. This shows that using detailed semantic descriptions helps the model better understand events and link audio-visual with text features. Without the $\mathcal{L}_{align}$, the ability of KDA to generalize worsened due to the lack of tight grouping within classes, leading to a decrease in HM by 52.5\%, 48.5\%, and 86.1\% on UCF-GZSL, VGGSound-GZSL, and ActivityNet-GZSL respectively. Without the $\mathcal{L}_{kaml}$, generalizability also worsened due to the lack of clear separations between classes, resulting in decreases in HM of KDA by 4.0\%, 8.7\%, and 26.3\% on these datasets respectively. These results show that the best performance of our model comes from using LLM-generated descriptions, $\mathcal{L}_{align}$ for intra-class compactness, and $\mathcal{L}_{kaml}$ for inter-class separability.

\noindent
\textbf{Influence of different rate $\lambda$.}
We conducted experiments to analyze different ratio coefficients $\lambda$ to balance the learning of inter-class and intra-class distributions. The experimental results are shown in Table~\ref{lamda}.
From the table, it can be observed that when $\lambda=10$, the model achieves the highest performance on UCF-GZSL and ActivityNet-GZSL. This indicates that at this specific value of $\lambda$, there is a good trade-off between intra-class and inter-class distribution learning. This finding suggests that selecting the appropriate $\lambda$ value is crucial for our model to achieve optimal performance.

\begin{table}[t]
    \centering
    \caption{Ablation study: Influence of different $\lambda$.}
    \label{lamda}
    \resizebox{\columnwidth}{!}{
    \begin{tabular}{ccccccccc}
    \toprule
       \multirow{2}{*}{Model} & \multicolumn{4}{c}{UCF-GZSL} & \multicolumn{4}{c}{ActivityNet-GZSL} \\ \cline{2-9}&S&U&HM&ZSL&S&U&HM&ZSL\\
       \midrule
       0& 45.19&17.17&24.88&20.10&2.75&3.65&3.13&4.80\\
       0.1&76.86&34.21&47.35&36.09&36.36  &14.98&21.21&16.65\\
       1&92.45&34.21&49.93&36.76&35.36&15.18&21.24&16.47\\
       5&87.62&34.76&49.79&39.08&42.05&14.86&21.96&17.07\\
       10&91.99&36.67&\textbf{52.44}&\textbf{39.06}&37.95& 15.95&\textbf{22.46}&\textbf{17.23}\\
       20&92.76&36.19&52.06&38.68&34.87&15.64&21.60&17.13\\
        \bottomrule
    \end{tabular}}
\end{table}

\begin{table}[t]
    \centering
    \caption{Ablation study: Influence of different text encoder.}
    \label{text}
    \resizebox{\columnwidth}{!}{
    \begin{tabular}{ccccccccc}
    \toprule
       \multirow{2}{*}{Model} & \multicolumn{4}{c}{UCF-GZSL} & \multicolumn{4}{c}{ActivityNet-GZSL} \\ \cline{2-9}&S&U&HM&ZSL&S&U&HM&ZSL\\
       \midrule
       Bert&36.54&13.21&19.40&14.24&19.55&6.43&9.67&7.84\\
       GPT&79.42&26.43&39.66&31.32&33.64&11.66&17.31&12.62\\
       Instructor&85.84&33.86&48.56&35.49&36.32& 15.10&21.33&15.96\\
       CLAP&84.66&32.63&47.10&34.75&35.98&15.38& 21.54 & 16.74\\
       CLIP&91.99&36.67&\textbf{52.44}&\textbf{39.06}&37.95&15.95&\textbf{22.46}&\textbf{17.23}\\
        \bottomrule
    \end{tabular}}
\end{table}

\noindent
\textbf{Influence of text encoder.}
In our study, we have considered several text encoders for our framework. These include BERT~\cite{devlin2018bert}, GPT-3~\cite{yang2022empirical}, Instructor~\cite{instructor}, CLAP~\cite{clap}, and CLIP~\cite{clip}. The performance of these encoders on the UCF-GZSL and ActivityNet-GZSL datasets is presented in Table~\ref{text}. From the table, we can observe that CLIP exhibits the highest performance. This is mainly due to the inherent ability of CLIP to align images and text, which is beneficial for our audio-visual zero-shot learning task. The alignment between images and text representations in CLIP aids in a more comprehensive cross-modal understanding and alignment, thereby leading to improved recognition of unseen event classes.

\noindent
\textbf{Evaluating different modalities.}
In Table~\ref{modality}, we compared our multi-modal KDA model with training our architecture using only unimodal inputs. In this case, we excluded the cross-modal attention block and trained each unimodal branch separately. The visual branch outperformed the audio branch, achieving a GZSL performance (HM) of 41.25\% compared to 18.68\% on the UCF-GZSL dataset. A similar trend was observed for the ZSL performance, with the visual branch achieving 33.78\% compared to 11.88\% for the audio branch. This pattern was also seen on the VGGSound-GZSL datasets, suggesting that visual input features contain more richer and comprehensive information about video content than audio inputs. This supports the idea that incorporating complementary information from both audio and visual inputs is highly beneficial for generalized zero-shot learning and zero-shot learning in video classification.

\subsection{Limitations}
In our proposed KnowleDge-Augmented audio-visual learning framework, we leverage the power of large language models to enhance the learning process. KDA demonstrates superior performance across all datasets, surpassing the current state-of-the-art methods. However, owever, KDA operates on time-averaged audio-visual input information, which may overlook fine semantic details. In addition, KDA uses the same description for all videos of the same class, which might introduce bias.

\begin{table}[t]
    \centering
    \caption{Ablation study: Influence of training KDA with different modalities.}
    \label{modality}
    \resizebox{\columnwidth}{!}{
    \begin{tabular}{ccccccccc}
    \toprule
       \multirow{2}{*}{Model} & \multicolumn{4}{c}{UCF-GZSL} & \multicolumn{4}{c}{VGGSound-GZSL} \\ \cline{2-9}&S&U&HM&ZSL&S&U&HM&ZSL\\
       \midrule
       Visual only&90.38&26.73&41.25& 33.78&18.32&6.61&9.71& 7.30\\
       Audio only&60.14&11.06&18.68&11.88&11.89&4.29&6.31&5.04\\
       KDA&91.99&36.67&\textbf{52.44}&\textbf{39.06}&24.10&7.22&\textbf{11.11}&\textbf{8.43}\\
        \bottomrule
    \end{tabular}}
\end{table}

\section{Conclusion}
In this paper, we introduce a novel Knowledge-Augmented audio-visual learning (KDA) framework for audio-visual zero-shot learning. This framework is designed to create a unified space for audio-visual and textual representations. By incorporating extra knowledge from large language models (LLMs) and utilizing a knowledge-aware adaptive margin loss, our model effectively tackles the challenge of heterogeneous feature representations and bridges the gap between audio-visual and semantic domains. Our experimental results demonstrate that KDA consistently outperforms current state-of-the-art methods on three benchmark event recognition datasets. These results emphasize the effectiveness of our approach in managing the complexities of audio-visual zero-shot learning tasks.

{
    \small
    \bibliographystyle{ieeenat_fullname}
    \bibliography{main}
}


\end{document}